\documentclass[11pt]{article}

\usepackage[margin=1in]{geometry}
\usepackage[table]{xcolor}
\usepackage{graphicx}
\usepackage{hyperref}

\usepackage[utf8]{inputenc}
\usepackage[strict]{csquotes}

\usepackage[USenglish]{babel}
\usepackage[font=footnotesize]{caption}
\usepackage[parfill]{parskip}
\usepackage{authblk}
\usepackage{booktabs}
\usepackage{microtype}
\usepackage{multirow}
\usepackage{subcaption}
\usepackage{tabularray}

\usepackage{algorithmicx}
\usepackage{algorithm}
\usepackage{algpseudocode}

\usepackage[defernumbers=true,maxbibnames=6,minbibnames=3,sorting=none]{biblatex}

\makeatletter
\let\blx@rerun@biber\relax
\makeatother

\definecolor{tabc0}{HTML}{f9ece4}
\definecolor{tabc1}{HTML}{cceff8}

\UseTblrLibrary{booktabs}
\hypersetup{
    colorlinks  = true,
    urlcolor    = black!32!blue,
    linkcolor   = black!32!blue,
    citecolor   = black!32!blue,
    pdfauthor   = {David Oniani, Xizhi Wu, Shyam Visweswaran, Sumit Kapoor, Shravan Kooragayalu, Katelyn Polanska, Yanshan Wang},
    pdftitle    = {Enhancing Large Language Models for Clinical Decision Support by Incorporating Clinical Practice Guidelines},
    pdfkeywords = {artificial intelligence, large language models, in-context learning, clinical decision support, clinical practice guidelines, covid-19},
    pdflang     = English,
    pdfpagemode = UseNone,
    pdfsubject  = {Enhancing Large Language Models for Clinical Decision Support by Incorporating Clinical Practice Guidelines}
}

\title{\textbf{Enhancing Large Language Models for Clinical Decision Support by Incorporating Clinical Practice Guidelines}}

\author[1,*]{David Oniani}
\author[1,\thanks{Equal contribution}]{Xizhi Wu}
\author[2,3,4]{Shyam Visweswaran}
\author[5]{Sumit Kapoor}
\author[6]{Shravan Kooragayalu}
\author[1]{Katelyn Polanska}
\author[1,2,3,4,\thanks{Corresponding author: \texttt{yanshan.wang@pitt.edu}}]{Yanshan Wang}

\affil[1]{Department of Health Information Management, University of Pittsburgh, Pittsburgh, PA, USA}
\affil[2]{Department of Biomedical Informatics, University of Pittsburgh, Pittsburgh, PA, USA}
\affil[3]{Intelligent Systems Program, University of Pittsburgh, Pittsburgh, PA, USA}
\affil[4]{Clinical and Translational Science Institute, University of Pittsburgh, Pittsburgh, PA, USA}
\affil[5]{Department of Critical Care Medicine, University of Pittsburgh Medical Center, Pittsburgh, PA, USA}
\affil[6]{Department of Pulmonary Medicine, University of Pittsburgh Medical Center Western Maryland, Cumberland, MD, USA}

\date{}

\begin{document}

\maketitle



\vspace{-32px}
\section*{Abstract}

\subsection*{Background}

Large~Language~Models~(LLMs), enhanced with Clinical~Practice~Guidelines~(CPGs), can significantly
improve Clinical~Decision~Support~(CDS). However, methods for incorporating CPGs into LLMs are not
well studied.

\subsection*{Methods}

We develop three distinct methods for incorporating CPGs into LLMs: Binary~Decision~Tree~(BDT),
Program-Aided~Graph~Construction~(PAGC), and Chain-of-Thought-Few-Shot~Prompting~(CoT-FSP). To
evaluate the effectiveness of the proposed methods, we create a set of synthetic patient
descriptions and conduct both automatic and human evaluation of the responses generated by four
LLMs: GPT-4, GPT-3.5~Turbo, LLaMA, and PaLM~2. Zero-Shot~Prompting~(ZSP) was used as the baseline
method. We focus on CDS for COVID-19 outpatient treatment as the case study.

\subsection*{Results}

All four LLMs exhibit improved performance when enhanced with CPGs compared to the baseline ZSP. BDT
outperformed both CoT-FSP and PAGC in automatic evaluation. All of the proposed methods demonstrated
high performance in human evaluation.

\subsection*{Conclusion}

LLMs enhanced with CPGs demonstrate superior performance, as compared to plain LLMs with ZSP, in
providing accurate recommendations for COVID-19 outpatient treatment, which also highlights the
potential for broader applications beyond the case study.


\section*{Introduction}

Large Language Models (LLMs)\footnote{Note that in this paper, the term \enquote{LLM} refers to a
\textit{generative}~Large~Language~Model~(LLM).} have opened new opportunities in many fields, with
applications ranging from creative~writing~\cite{ann2022} to chemical
research~\cite{boiko2023}. LLMs have also been studied in the context of healthcare, where they
have proven successful at numerous tasks, including clinical reasoning~\cite{strong2023},
question-answering~\cite{lambert2024}, and medical evidence summarization~\cite{tang2023},
among other applications. Moreover, recent research produced powerful healthcare LLMs, such as
GatorTron~\cite{yang2022} and Med-PaLM~\cite{singhal2023}.

The success and popularity of LLMs are at least partially due to their In-Context Learning (ICL)
capability~\cite{brown2023}. ICL is a learning method that does not require model parameter
updates but \enquote{learns} from the given context. It can be thought of as a higher-order function
that takes several few-shot examples (i.e., the context) to produce the predictor
function~\cite{garg2022}. The emergence of ICL-capable LLMs has provided an alternative to a
typical pre-train, fine-tune, and predict pipeline, with prompting replacing fine-tuning. Prompting
is a novel paradigm where, with the use of textual prompts, downstream tasks are modeled as those
typically solved during pre-training~\cite{liu2023}. It has given rise to
Prompt~Engineering~(PE)~\cite{reynolds2021}, a field that aims to design prompts that enhance
ICL and improve LLM reasoning. One of several widely used PE
methods~\cite{ouyang2022,wang2023,yao2023} is Chain-of-Thought~(CoT)~\cite{wei2022}
prompting, which attempts to improve LLM reasoning by providing intermediate reasoning steps as part
of the prompt.

Clinical~Decision~Support~(CDS) supports healthcare decision-making by enabling the timely delivery
of evidence-based guidelines at the point of care~\cite{dullabh2022,sutton2020}. Optimizing CDS
has the potential to significantly improve clinical workflows, and it has been one of the major
topics of discussion in healthcare~\cite{osheroff2007}. The importance of CDS, paired with its
recent focus on patient-centered solutions~\cite{dullabh2022_2,sittig2023}, demands novel
methods for improving the quality of healthcare decision-making, ultimately facilitating better
health outcomes. At the same time, the literature on LLM-driven algorithmic methods that incorporate
CPGs is limited. LLMs offer many benefits for CDS, such as timely decision support, ease of use due
to their interactive nature, and the ability to capture both Clinical~Practice~Guidelines~(CPGs) and
patient-specific information.

In this paper, we propose three methods for improving LLMs for CDS by incorporating CPGs:
Binary~Decision~Tree~(BDT), Program-Aided~Graph~Construction~(PAGC), and
Chain-of-Thought-Few-Shot~Prompting~(CoT-FSP). Specifically, we enhance LLMs by incorporating CPGs
and evaluate the performance of the proposed approaches on synthetic patient descriptions. The
evaluation was done by two physicians from the University~of~Pittsburgh Medical~Center~(UPMC). We
used Zero-Shot~Prompting~(ZSP) as the baseline. The experimental results show that the proposed
methods provide significant improvements over the baseline, which speaks to the effectiveness of the
methods.


\subsection*{Related Work}

The field of using LLMs for CDS is rapidly evolving. Here we summarized related work up to the point
when this paper was drafted. A recent study considered ChatGPT~\cite{chatgpt} LLM for improving
CDS~\cite{liu2023_2}, reaching a conclusion that LLMs can be important complementary parts for
optimizing CDS. In another study, researchers examined LLMs for improving CDS in personalized
oncology~\cite{benary2023}, reaching a similar conclusion that LLMs did not reach the quality
and credibility of human experts, but generated helpful responses that could complement established
procedures. There have also been other studies evaluating LLMs for CDS in specific fields, such as
cardiology~\cite{lee2023} and orthopedics~\cite{chatterjee2023}. It should be noted that
none of the studies augmented LLMs with CPGs, nor did they consider advanced prompting methods such
as the ones proposed in our study. To the best of our knowledge, our study is the first one to
develop and evaluate such methods.


\section*{Methods}

\begin{figure}
    \includegraphics[width=\linewidth]{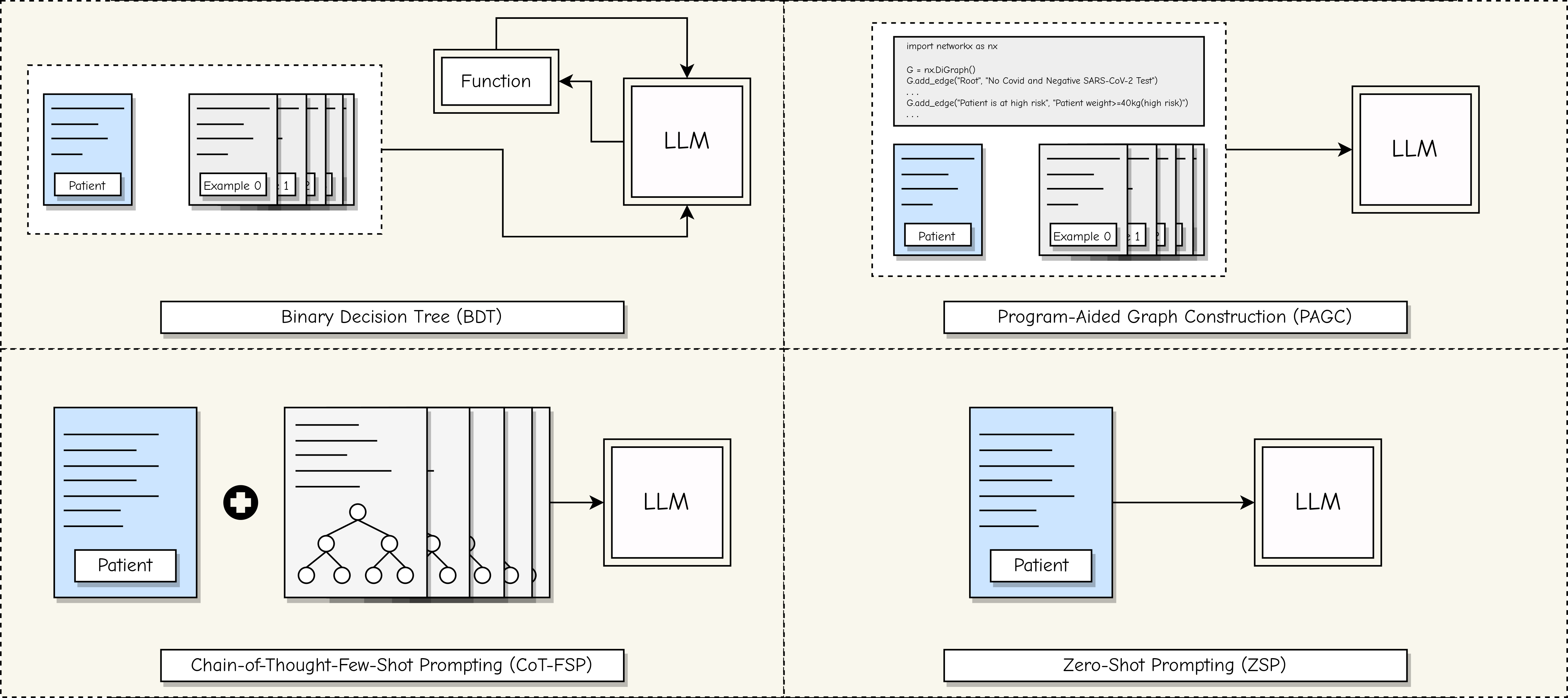}
    \caption{The figure shows the three proposed methods and the baseline Zero-Shot~Prompting~(ZSP)
             method. In the case of the Binary~Decision~Tree~(BDT) method, we use a recursive
             function to call the LLM with prompts. For Program-Aided~Graph~Construction~(PAGC), a
             program is a part of the prompt passed to the LLM.
             Chain-of-Thought-Few-Shot~Prompting~(CoT-FSP) uses several few-shot examples for
             guiding the LLM. Finally, ZSP only takes the patient description to produce the result.
             \textit{Note that for BDT, PAGC, and CoT-FSP, the prompt typically contains a task
             description, patient description, and several few-shot examples besides additions
             specific to a method (e.g., a program in the case of PAGC)}.}
    \label{fig:methods}
\end{figure}

We propose three methods enhanced by CPGs as well as a baseline Zero-Shot~Prompting~(ZSP) method.
Figure~\ref{fig:methods} illustrates the methods.

\subsection*{COVID-19 Treatment Guidelines}

We focused on supporting outpatient treatment decisions for Coronavirus Disease 2019 (COVID-19).
Because COVID-19 was a novel disease at the start of the pandemic, clinical practice guidelines
evolved rapidly in response to new discoveries and treatments. Keeping up with the frequent
guideline updates was difficult for healthcare providers. In such a situation, a CDS tool like LLM
can be valuable in providing reliable support for decision making in clinical care. For the COVID-19
outpatient treatment guidelines, we used the Centers for Disease Control and Prevention (CDC) and
Infectious Diseases Society of America (IDSA) COVID-19 Outpatient Treatment
Guidelines~\cite{cpgs}. The CPGs provide a comprehensive step-by-step approach that outlines
outpatient treatment options. We slightly revised the guidelines to reflect insights and
recommendations from our physicians. Body weight, in particular, was used as a key determinant for
COVID-19 treatment since many medication dosages are weight-based. We considered the patient's
underlying medical conditions, COVID-19 related symptoms, and the duration of symptoms to determine
if they were in the high-risk versus low-risk category. Based on this information, the guidelines
offer three outpatient treatment options for high-risk patients: \textit{Paxlovid},
\textit{Remdesivir}, or \textit{Molnupiravir}, and one treatment suggestion for low-risk patients:
\textit{Supportive Care}. The revised COVID-19 Outpatient Treatment Guidelines were used to build
LLM-enhanced CDS systems for COVID-19 outpatient treatment decision support and can be found in the
\nameref{appendix}.

\subsection*{Binary Decision Tree (BDT)}

Binary~Decision~Tree~(BDT) is a recursive algorithm guided by CPGs. We model CPGs as a binary tree
and use an LLM to navigate the tree. There are two calls to the LLM. In the first call, an input to
an LLM contains a description of the task, several few-shot examples, and a question from CPGs
(i.e., node value). The generated output is stored. In the second call, we use the response from the
first call to ask a yes-no question of whether the answer was affirmative or negative. Depending on
the generated response, we recurse to the left or right subtree till reaching one of the leaf nodes.
Algorithm~\ref{bdt} is the pseudocode for the method.

\begin{algorithm}
    \caption{Binary~Decision~Tree~(BDT) algorithm}
    \label{bdt}
    \begin{algorithmic}[1]
        \Require \(P_1\): task description, patient description, and several few-shot examples
        \Require \(P_2\): task description and several few-shot examples for YES/NO question
        \Require \(M\): pre-initialized Large Language Model (LLM)
        \Procedure{BDT}{node, responses\(=\)()}
            \If{\textbf{not} (\textnormal{node.left} \textbf{or} node.right)} \Comment{Base case}
                \State \textbf{return} (node.value, responses)
            \EndIf
            \State \(O_1 \gets M(P_1\) \(+\) node.question) \Comment{First LLM call}
            \State responses.add(\(O_1\))
            \State \(O_2 \gets M(\textnormal{\enquote{Response YES or NO?}} + P_2 +\) node.question \(+ O_1)\) \Comment{Second LLM call}
            \If{\(O_2 = \textnormal{\enquote{YES}}\)} \Comment{Recursive step: recurse to left or right subtree}
                \State (result, responses) = BDT(node.left, responses)
            \ElsIf{\(O_2 = \textnormal{\enquote{NO}}\)}
                \State (result, responses) = BDT(node.right, responses)
            \Else
                \State \textbf{throw Exception}
            \EndIf
            \State \textbf{return} (result, responses)
        \EndProcedure
    \end{algorithmic}
\end{algorithm}

\subsection*{Chain-of-Thought-Few-Shot~Prompting~(CoT-FSP)}

In this method, we use a prompt to describe the CPGs. Specifically, we construct a prompt containing
a description of the task, five few-shot examples, and an if-else description of CPGs.

The if-else description is an algorithmic seven-step CoT prompt. At every step, CoT-FSP decides
between selecting a specific treatment and progressing to the next step. The selection of a specific
treatment ends the chain. In the first step, the algorithm checks whether the patient has COVID-19,
and if they do, it moves to the second step. Otherwise, it provides the treatment suggestion
\enquote{Vaccination and booster is recommended}. In the second step, the algorithm checks whether
the patient needs hospitalization or increased oxygen. In the third step, the algorithm looks for
risk factors for severe COVID-19. If the hypothetical patient has risk factors, it goes to the
fourth, fifth, and seventh steps to find an appropriate treatment such as Paxlovid, Remdesivir, or
Molnupiravir. If none of the treatment options are available for the patient, it will suggest the
patient get \enquote{monitoring and supportive care} and end the process.

The five few-shot examples are five hypothetical patient descriptions and five answers on providing
treatment suggestions after going through the step-by-step if-else process. We carefully selected
more complex patient cases to improve ICL and handle challenging scenarios. Algorithm~\ref{cotfsp}
shows the pseudocode for the method.

\begin{algorithm}
    \caption{Chain-of-Thought-Few-Shot~Prompting~(CoT-FSP) algorithm}
    \label{cotfsp}
    \begin{algorithmic}[1]
        \Require \(P_1\): patient description, task description, and several CPG-enhanced few-shot examples
        \Require \(P_2\): text description of the tree via if-else statements
        \Require \(M\): pre-initialized Large Language Model (LLM)
        \Procedure{CoT-FSP}{}
            \State \(P \gets P_1 + P_2\)
            \State \(O \gets M(P)\)
            \State \textbf{return} \(O\)
        \EndProcedure
    \end{algorithmic}
\end{algorithm}

\subsection*{Program-Aided Graph Construction (PAGC)}

Inspired by Program-aided Language Models (PAL)~\cite{gao2023}, the PAGC approach defines a
\texttt{networkx}\footnote{\url{https://networkx.org/}} graph directly in the prompt and uses five
few-shot examples to select candidates.

The \texttt{networkx} graph represents the revised COVID-19 CPGs, with eight leaf nodes representing
eight COVID-19 outpatient treatment suggestions and each internal node representing the patient's
possible medical condition or characteristics. The method is accompanied by a candidate selection
algorithm that selects candidate nodes in the \texttt{networkx} graph that match the patient
description. After selecting all the candidate nodes, a path to one of the leaf nodes is created.

The five few-shot examples are designed to be challenging, improving the LLM's ability to handle
complex cases. Algorithm~\ref{pagc} illustrates the pseudocode for the method.

\begin{algorithm}
    \caption{Program-Aided Graph Construction (PAGC) algorithm}
    \label{pagc}
    \begin{algorithmic}[1]
        \Require $P$: task description and patient description
        \Require $C$: code description of the algorithm
        \Require $M$: pre-initialized Large Language Model (LLM)
        \Procedure{PAGC}{}
            \State $O \gets LLM(P + C)$
            \State \textbf{return} $O$
        \EndProcedure
    \end{algorithmic}
\end{algorithm}

\subsection*{Zero-Shot Prompting (ZSP)}

Zero-Shot~Prompting~(ZSP) method constructs a prompt that contains a task description and patient
description, followed by a query. In contrast with BDT, CoT-FSP, and PAGC methods, ZSP does not
utilize CPGs or few-shot examples. Instead, ZSP produces the output directly from a patient
description. Algorithm~\ref{zsp} describes this method.

\begin{algorithm}
    \caption{Zero-Shot Prompting (ZSP) algorithm}
    \label{zsp}
    \begin{algorithmic}[1]
        \Require $P$: task description and patient description
        \Require $M$: pre-initialized Large Language Model (LLM)
        \Procedure{ZSP}{}
            \State $O \gets LLM(P)$
            \State \textbf{return} $O$
        \EndProcedure
    \end{algorithmic}
\end{algorithm}

\subsection*{Synthetic Patient Dataset}

For rigorous evaluation of the proposed methods, we created a synthetic patient dataset based on
CPGs. We modeled CPGs as a binary tree with eight leaf nodes. There were thirteen different paths to
reach one of the leaves. Considering three different difficulty categories (\textit{easy},
\textit{medium}, and \textit{hard}), we created \(3 \times 13 = 39\) synthetic patients to examine
all possible paths. All the synthetic patients, including their \textit{difficulty}, are listed in
the \nameref{appendix}.

In the \textit{easy} category, patient descriptions adhered closely to the phrases used in the
COVID-19 CGPs. Such an approach allowed for assessing methods in making correct decisions at each
checkpoint and identifying the correct treatment path following the guidelines.

The \textit{medium} category contained patient descriptions that used synonyms and phrases
semantically equivalent to those in the CPGs. For example, instead of describing a patient as one
\enquote{who had a positive COVID-19 test}, the description mentioned \enquote{a positive Polymerase
Chain Reaction (PCR) test}. The goal was to evaluate the ability to comprehend the semantic meaning
in the patient description, correctly map it to the relevant checkpoint, and make the correct
decision.

The \textit{hard} category included patient descriptions that differed considerably from the phrases
used in the CPGs. We also intentionally included unrelated and subtle information. Hence, the
methods would not only need to extract symptoms from the description but also filter out irrelevant
information.

\subsection*{Evaluation}

The evaluation has two stages: best method selection and human annotation. We considered four LLMs,
GPT-4~\cite{openai2023gpt4}, GPT-3.5 Turbo~\cite{openaigpt3.5turbo},
LLaMA~\cite{touvron2023}, and PaLM~2~\cite{anil2023}, for a more holistic evaluation.

In the method selection stage, we automatically evaluate the approaches and select those with an
F-score greater than 0.5. Systematically evaluating few-shot classification performance is
challenging as predictions on small datasets can be unstable~\cite{gao2021}. To ensure the
reproducibility and robustness of results, we performed four runs with random seeds (\texttt{9631},
\texttt{4603}, \texttt{6367}, and \texttt{4057}) and computed the mean for each metric\footnote{Note
that PaLM~2 does not provide a way to set the seed, and we just re-ran the model several times.
Besides, GPT-based models do not always produce the same output when called with the same random
seed, which is a known issue:
\url{https://community.openai.com/t/seed-param-and-reproducible-output-do-not-work/487245}.}.

The human evaluation has two rounds: a round for ensuring agreement among two UPMC physicians
(authors S.K. and S.K.) and the final LLM response annotation round\footnote{Note that model names (e.g., GPT-4, PaLM~2, etc.) were hidden from annotators to ensure fair and unbiased evaluation.}. We used random seed \texttt{8747}
for generating the responses. In the first round, the annotators rated ten questions for evaluation.
We used Gwet's AC1 for computing Inter-Annotator~Agreement~(IAA)~\cite{wongpakaran2013}. For
interpreting IAA scores, we used Landis~and~Koch~scale~\cite{landis1977}.

\begin{table}[htbp]
  \centering
  \footnotesize
  \begin{tblr}{
    colspec = {|[1pt,black]lcl|[1pt,black]},
    row{odd} = {tabc0},
    row{even} = {white},
    row{1} = {tabc1}
  }
    \hline[1pt,black]
    \textbf{Table~\ref{tab:ac1}: Gwet's AC1 by question for the first round.\(^\star\)}&\\
    \hline[1pt,black]
    \textbf{Question}                   & \textbf{Score} & \textbf{Interpretation}\\
    Presence of Incorrect Medical Content & 0.87         & Almost Perfect\\
    Omission of Content                   & 0.60         & Moderate\\
    Presence of Possible Harmful Content  & 0.77         & Substantial\\
    \hline[1pt,black]
  \end{tblr}
  \caption{\(^\star\) Gwet's AC1 scores for three different questions asked to the physicians.
           Landis and Koch interpretations for the scores are also included. \enquote{Omission of
           Content} is the only category with \textit{Moderate} agreement. It should also be noted
           that even in this case, the score signifies borderline substantial agreement (i.e., would
           have been substantial if the obtained score was \(0.61\)).}
  \label{tab:ac1}
\end{table}

For physician annotations, we have three different evaluation categories: \textit{presence of
incorrect medical content}, \textit{ommision of content}, and \textit{presence of possible harmful
content}~\cite{singhal2023}. We designed these categories to comprehensively capture and
represent the errors and mistakes we found in our testing process. The human evaluation scale
ranges from 0 to 2 for all three question categories and is described in detail in the
\nameref{appendix}.

The \textit{presence of incorrect medical content} category aims to discern the accuracy of the
generated answers concerning medical content. The objective is to determine whether the responses
contain information conflicting with established medical guidance. The \textit{omission of content}
evaluates the completeness of the generated answers and aims to identify instances where essential
information is either inadequately addressed or omitted entirely. Understanding the model's ability
to provide comprehensive responses is crucial for its practical utility in clinical decision-making.
The \textit{presence of possible harmful content} is an essential consideration in our evaluation
framework as the identification of potential harm that LLMs could pose to users relying on the
generated answers. This category is responsible for ensuring the safety and ethical use of the
methods in real-world applications, particularly in sensitive domains such as healthcare.

Table~\ref{tab:ac1} shows the results for the first round of annotations. Since the score for
\enquote{omission of content} was \textit{Moderate} per Landis~and~Koch~scale, we held a meeting
where the physicians discussed discrepancies, reaching the perfect agreement on the annotations for
this specific category. We note that \(0.61\) would have already been a substantial agreement, but
we got \(0.60\), which was \(0.01\) less.


\section*{Results}

\begin{table}[htbp]
  \centering
  \footnotesize
  \begin{tblr}{
    colspec = {|[1pt,black]lcccc|[1pt,black]},
    row{odd} = {tabc0},
    row{even} = {white},
    row{1} = {tabc1}
  }
    \hline[1pt,black]
    \textbf{Table~\ref{tab:method_selection}: F-scores for the methods.\(^\star\)} & & & &\\
    \hline[1pt,black]
    \textbf{LLM}  & \textbf{BDT} & \textbf{CoT-FSP} & \textbf{PAGC} & \textbf{ZSP}\\
    GPT-4         & 1.00\(^1\)   & 0.97\(^2\)       & 0.83\(^4\)    & 0.47\\
    GPT-3.5~Turbo & 0.85\(^3\)   & 0.69\(^6\)       & 0.38          & 0.26\\
    LLaMA-13b     & 0.37         & 0.31             & 0.42          & 0.31\\
    PaLM~2        & 0.71\(^5\)   & 0.58\(^7\)       & 0.41          & 0.01\\
    \hline[1pt,black]
  \end{tblr}
  \caption{\(^\star\) F-score metrics averaged over four different runs with four randomly generated
           four-digit prime random seeds: \texttt{9631}, \texttt{4603}, \texttt{6367},
           \texttt{4057}. The selection of methods based on F-score was done using this table. Only
           those with F-score \(>\) 0.5 were selected. Selected methods have a rank number as
           superscript.}
  \label{tab:method_selection}
\end{table}

We first performed the automatic evaluation and then picked the seven most performant methods for
the human evaluation stage, namely \textit{GPT-4~BDT}, \textit{GPT-4~CoT-FSP},
\textit{GPT-3.5~Turbo~BDT}, \textit{GPT-4~PAGC}, \textit{PaLM~2~BDT},
\textit{GPT-3.5~Turbo~CoT-FSP}, and \textit{PaLM~2~CoT-FSP}.

Pairing our methods with LLaMA-13b did not yield promising results. As a result, the model was
excluded entirely from the human evaluation stage. In its defense, however, LLaMA-13b is
significantly smaller than the other models, and a larger LLaMA model could perform better. ZSP
performed poorly across all models, and aw also excluded it due to this reason.
Table~\ref{tab:method_selection} shows and ranks the selected methods.

In the human evaluation stage, two UPMC physicians (authors S.K. and S.K.) annotated 49 responses
generated using the proposed methods. We generated seven responses using every method and computed
the mean score for each evaluation category. Table~\ref{tab:human_eval} shows the human evaluation
results.

\begin{table}[htbp]
  \centering
  \footnotesize
  \begin{tblr}{
    colspec = {|[1pt,black]lXXXXXXX|[1pt,black]},
    row{odd} = {tabc0},
    row{even} = {white},
    row{1} = {tabc1}
  }
    \hline[1pt,black]
    \textbf{Table~\ref{tab:human_eval}: Human evaluation scores.\(^\star\)} & & & & & & &\\
    \hline[1pt,black]
    \textbf{Category}                          & \textbf{GPT-4 BDT}     & \textbf{GPT-4 CoT-FSP}     & \textbf{GPT-3.5 Turbo BDT} & \textbf{GPT-4 PAGC} & \textbf{PaLM~2 BDT} & \textbf{GPT-3.5 Turbo CoT-FSP} & \textbf{PaLM~2 CoT-FSP}\\
    Presence of Incorrect Medical Content & 1.14                   & 2.00                       & 1.71                       & 2.00                & 2.00              & 1.14                           & 1.14\\
    Omission of Content                   & 0.86                   & 2.00                       & 2.00                       & 2.00                & 2.00              & 1.43                           & 2.00\\
    Presence of Possible Harmful Content  & 1.29                   & 2.00                       & 1.86                       & 2.00                & 2.00              & 1.14                           & 1.14\\
    Overall Average                       & 1.10                   & 2.00                       & 1.86                       & 2.00                & 2.00              & 1.24                           & 1.43\\
    \hline[1pt,black]
  \end{tblr}
  \caption{\(^\star\) Human evaluation results for the methods selected via automatic evaluation. We
           computed the average score for each evaluation category. GPT-4 CoT-FSP, GPT-4 PAGC, and
           PaLM~2 BDT obtained the perfect scores, with the highest quality of responses (rating of
           2) for all queries. GPT-3.5 Turbo BDT was next, with an overall average of 1.86. PaLM~2
           CoT-FSP, GPT-3.5 Turbo CoT-FSP, and GPT-4 BDT showed the worst performance, with average
           scores of 1.43, 1.24, and 1.10, respectively.}
  \label{tab:human_eval}
\end{table}

GPT-4 CoT-FSP, GPT-4 PAGC, and PaLM~2 BDT obtained the perfect scores, with the highest quality of
responses (rating of 2) for all queries. GPT-3.5 Turbo BDT was next, with an overall average of
1.86. PaLM~2 CoT-FSP, GPT-3.5 Turbo CoT-FSP, and GPT-4 BDT showed the worst performance, with
average scores of 1.43, 1.24, and 1.10, respectively.

We should note that in all cases, the average score was above 1.00, which speaks to the promising
performance of the approaches. Interestingly, the best method in the automatic evaluation, GPT-4
BDT, did not perform as well in human evaluation. At the same time, among models that utilized BDT,
PaLM~2 and GPT-3.5 Turbo demonstrated strong performance across all queries.


\section*{Discussion}

Aiming to improve CDS, we proposed three new methods for incorporating CPGs into LLMs. We took
COVID-19 CDS as the case study and created a set of synthetic patients. We first selected methods
with an F-score higher than \(0.5\) and then performed a rigorous human evaluation of the
approaches. The proposed methods outperformed the baseline ZSP and have shown high performance
across the tasks. Our work also opens future directions of research.

\begin{figure}[H]
    \centering
    \begin{subfigure}{0.5\linewidth}
        \centering
        \includegraphics[width=0.98\linewidth]{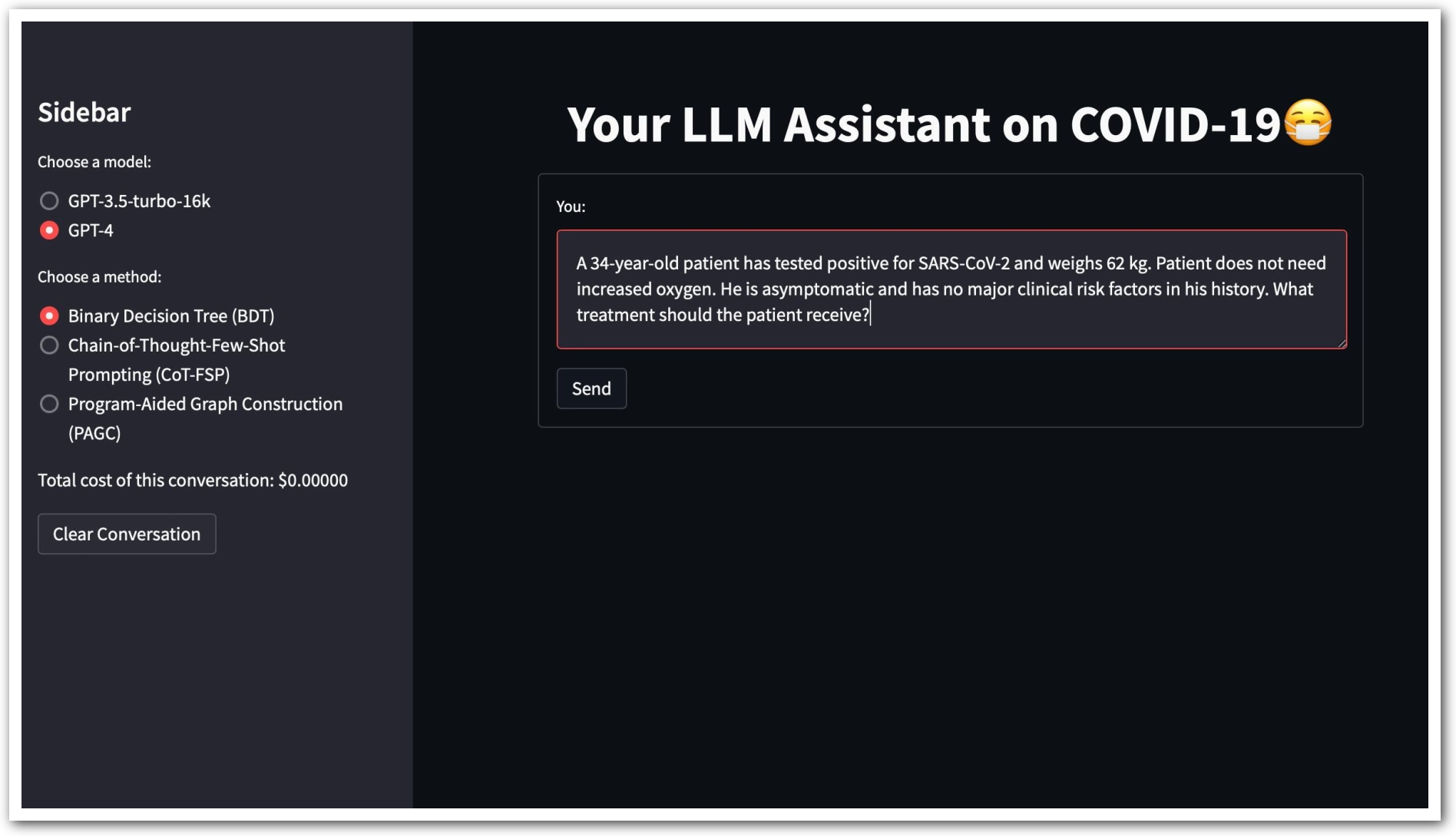}
        \caption{Prototype chatbot interface \textit{before} generating the response.}
        \label{fig:llmcovidbefore}
    \end{subfigure}%
    \begin{subfigure}{0.5\linewidth}
        \centering
        \includegraphics[width=0.98\linewidth]{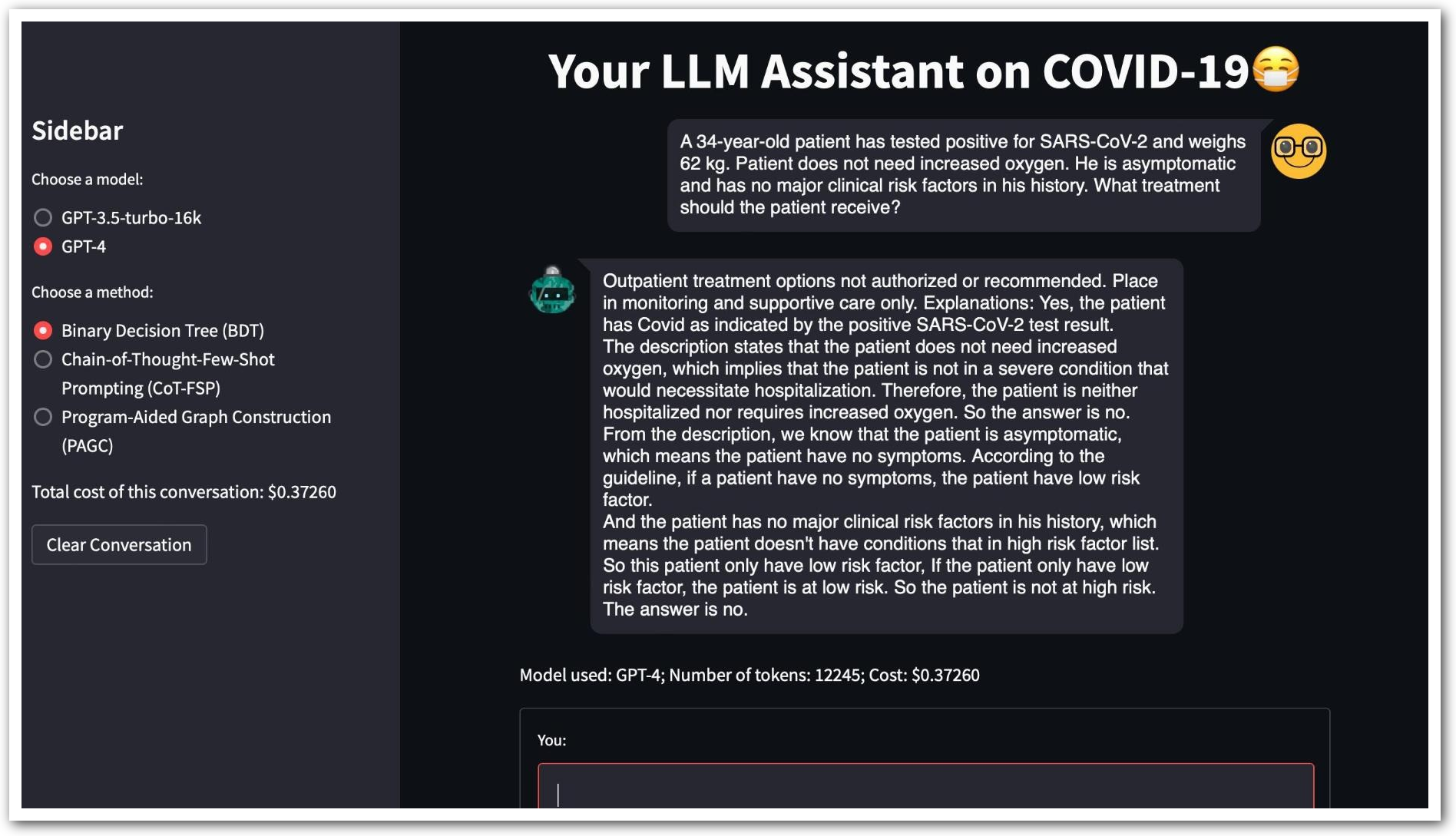}
        \caption{Prototype chatbot interface \textit{after} generating the response.}
        \label{fig:llmcovidafter}
    \end{subfigure}
    \caption{The figure shows the user interface of the chatbot system that implements the proposed
             methods. Figure~\ref{fig:llmcovidafter} shows the interface with the prompt but
             \textit{before} generating the response. Figure~\ref{fig:llmcovidbefore} shows the
             interface \textit{after} after generating the response. We developed the system as part
             of the research effort to demonstrate real-world implementation of the methods and
             collect user feedback.}
    \label{fig:llmcovid}
\end{figure}

The promising performance of the proposed methods allows for incorporating them as part of software
systems, such as chatbots, that streamline CDS. As part of this work, we have also built a
prototype, shown in Figure~\ref{fig:llmcovid}. The system takes the prompt from a medical
professional as input and provides treatment recommendations, enhancing the clinical pipeline and
facilitating accurate healthcare delivery. A case study exploring the systems utilizing the proposed
methods in real-world clinical settings is another interesting future direction of research.

Future research can also focus on improving the proposed methods. An example of such improvement
could be tree pruning or node skipping in the case of BDT. In other words, some responses may
contain enough information to answer several questions, allowing for shortcuts across the tree. The
proposition of completely different LLM-based approaches for enhancing CDS can also be another
avenue of exploration.

In addition, we would like to emphasize the need for transparent and privacy-aware development of
LLMs, which is especially important in the case of healthcare~\cite{oniani2023}, where even a
small-scale accidental leak of Protected Health Information (PHI) can be dangerous. Open-source
models offer transparency, allowing researchers to audit the codebase if necessary. On the other
hand, proprietary LLMs typically offer the Application Programming Interface (API) model, where the
codebase is not publicly available, and there is no clarity on where the user or user request data
is kept (and for how long). Among the four models used in our study, only
LLaMA\footnote{\url{https://github.com/facebookresearch/llama}} and
PaLM\footnote{\url{https://github.com/conceptofmind/PaLM}} were open-source.

Finally, we also note that experiments with LLMs can be costly. In our case, the associated costs
for conducting experiments with GPT-4 and GPT-3.5 Turbo were approximately \$500 and \$150,
respectively. Thus, we paid \$650 for two models alone, with GPT-4 being roughly 3.3 times more
expensive than GPT-3.5 Turbo. And this is with only half of the models requiring a payment. The
other two LLMs, LLaMA and PaLM 2, were free-to-use and open-source, with no expenses besides utility
costs for running a server. For LLaMA, we used our lab server with Quadro RTX 8000 GPUs. As for PaLM
2, we utilized the provided API\footnote{\url{https://ai.google.dev/models/palm}}.



\section*{Acknowledgements}

None.


\section*{Contributions}

D.O. led the study, designed the experiments, helped conduct the experiments, analyzed the results,
and wrote, reviewed, and revised the paper. X.W. designed and conducted the experiments, analyzed
the results, and wrote, reviewed, and revised the paper. S.V. conceptualized the study and revised
the paper. S.K. and S.K. annotated the results and reviewed the paper. K.P. analyzed the results and
reviewed the paper. Y.W. conceptualized the study and reviewed and revised the paper.


\printbibliography


\clearpage

\section*{Appendix} \label{appendix}

\subsection*{Human Evaluation Guidelines for COVID-19 Treatment Recommendation}

We designed three human evaluation categories with categorical ratings:

\begin{itemize}
    \item Presence of Incorrect Medical Content
        \begin{enumerate}
            \item Yes, great clinical significance (there is incorrect medical content and it will
                  make great clinical significance on the treatment recommendation, such as an
                  incorrect treatment suggestions based on patient's conditions, like recommending
                  patient who does not wish to go to hospital for Remdesivir infusion treatment)
            \item Yes, little clinical significance (there is incorrect medical content and it will
                  make little clinical significance on the treatment recommendation)
            \item No (i.e., no incorrect medical content)
        \end{enumerate}

    \item Omission of Content
        \begin{enumerate}
            \item Yes, great clinical significance (there is omission of the input patient's
                  condition or characteristics and it will make great clinical significance on the
                  treatment recommendation, such as changing the result of recommendation)
            \item Yes, little clinical significance (there is omission of the patient’s condition or
                  characteristics and it will make little clinical significance on the treatment
                  recommendation, for example, even though omission of content occurred, it does not
                  change the ultimate result)
            \item No (i.e., no omission of content)
        \end{enumerate}

    \item Presence of Possible Harmful Content
        \begin{enumerate}
            \item Death or severe harm
            \item Moderate or mild harm
            \item No harm (i.e., no harmful content)
        \end{enumerate}
\end{itemize}

\subsection*{Paper Graphic}

\begin{figure}[H]
    \centering
    \includegraphics[width=\linewidth]{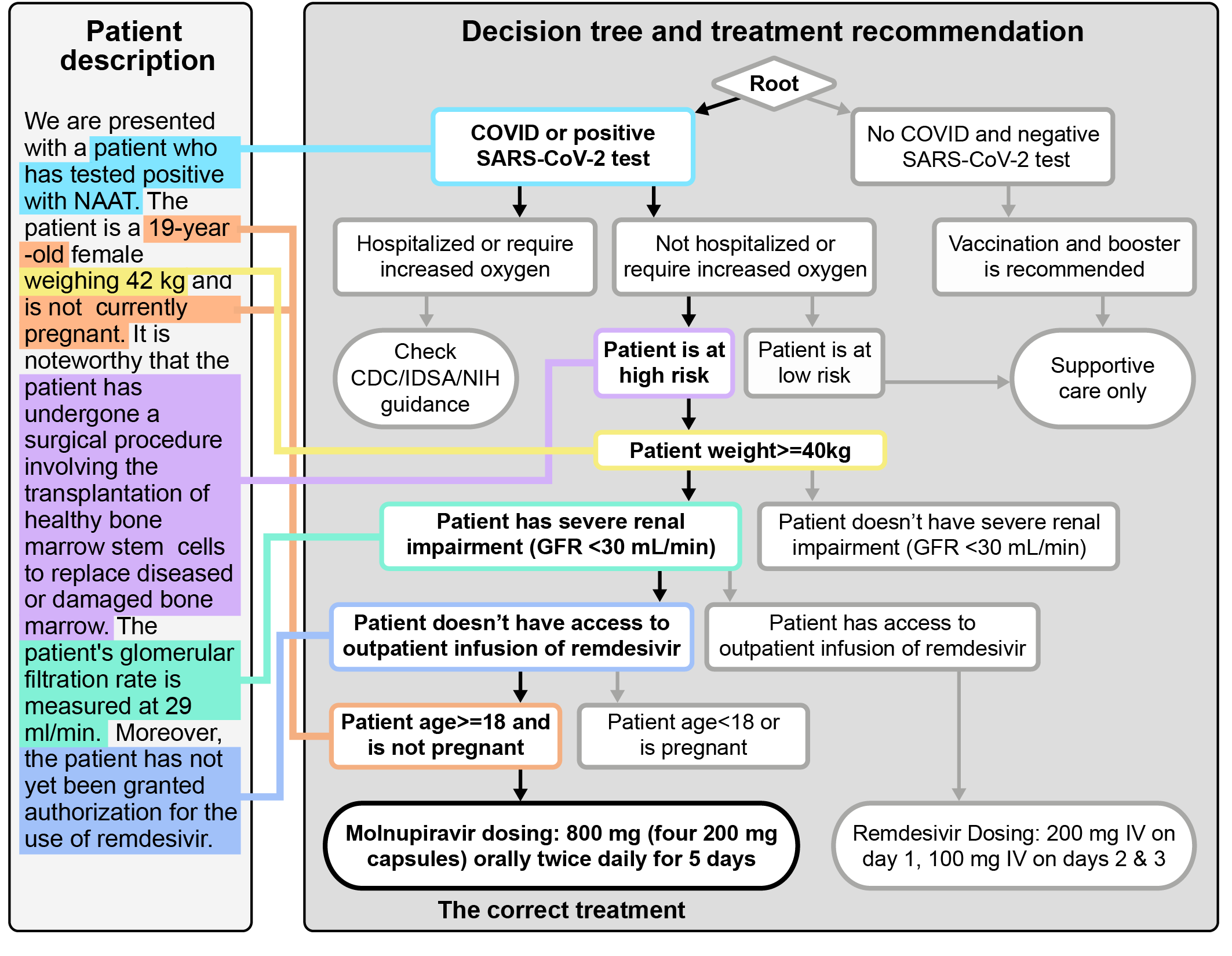}
    \caption{A graphic for the paper.}
\end{figure}

\subsection*{Clinical Practice Guidelines}

\begin{figure}[H]
    \centering
    \includegraphics[width=0.67\linewidth]{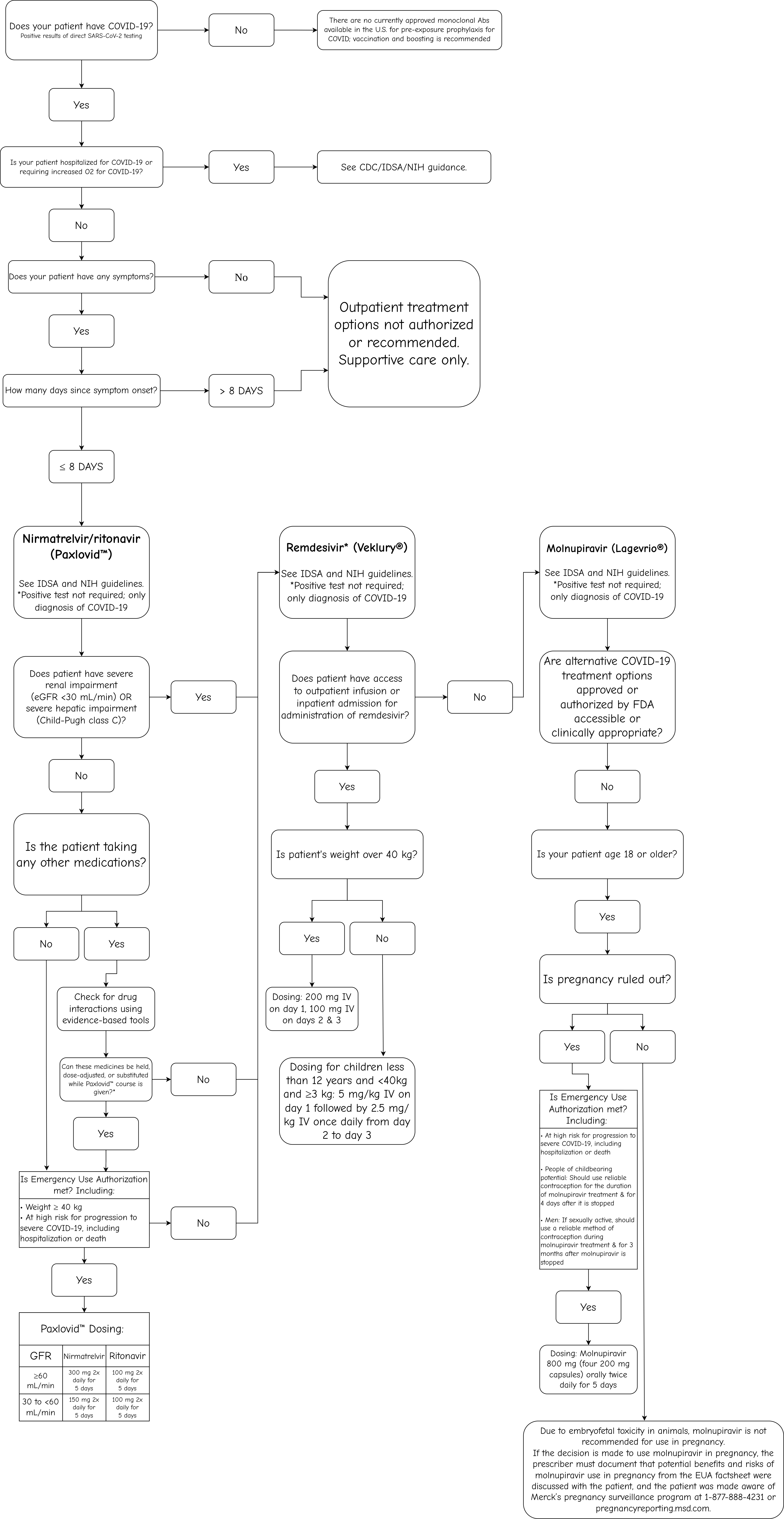}
    \caption{Clinical Practice Guidelines (CPGs) for COVID-19. We used modified the Centers for
             Disease Control and Prevention (CDC) and Infectious Diseases Society of America (IDSA)
             COVID-19 Outpatient Treatment Guidelines.}
\end{figure}

\subsection*{Synthetic Patients}

\DefTblrTemplate {caption-text}{short}{\InsertTblrText{entry}}

\DefTblrTemplate{capcont}{short}{%
  \par\centering
  \UseTblrTemplate {caption-tag}{default}%
  \UseTblrTemplate {caption-sep}{default}%
  \UseTblrTemplate {caption-text}{short}
  \UseTblrTemplate {conthead-text}{default}
  \par
}
\NewTblrTheme{shortcaption}{%
    \SetTblrTemplate{capcont}{short}%
}

\begin{longtblr}[theme = shortcaption, caption = {All 39 synthetic patients used in the study.}]{
  colspec = {|[1pt,black]X[2]X[1]X[1]|[1pt,black]},
  row{odd} = {tabc0},
  row{even} = {white},
  row{1} = {tabc1}
}
  \hline[1pt,black]
  \textbf{Table 1: All 39 synthetic patients used in the study.\(^\star\)} &\\
  \hline[1pt,black]
    \textbf{Synthetic Patient} & \textbf{Correct Answer} & \textbf{Difficulty}\\
    Patient is positive for covid-19, is hypoxic and needs supplemental oxygen. Patient has a history of renal transplant. His eGFR is 61 ml/min. He takes medications that interact with Paxlovid. He has no contraindications for remdesivir and can receive it at nearby infusion center. & Check CDC/IDSA/NIH Guidance & hard\\
    An 18-year-old patient has tested positive for Covid-19. He weighs 60 kg and/ asymptomatic. The patient does not need hospitalization. He has eGFR of 72 ml/min. He also takes medications that interact with Paxlovid. He cannot take remdesivir as he has no access to infusion center and does not want to be admitted to hospital. & Outpatient treatment options not authorized or recommended. Place in monitoring and supportive care only & hard\\
    An 18-year-old boy/male, weighing 60 kg tested for Covid-19 and results are negative. He has a history of genetic blood disorder and chronic kidney disease with eGFR of 32 ml/min. Patient takes medication that induces CYP3A4 enzyme and interacts with paxlovid. He cannot take remdesivir as he does not have nearby infusion center and does not want to be admitted to the hospital. & Vaccination and booster is recommended & hard\\
    A 28-year-old biological female tested positive for Covid-19. She weighs 65 kg. She is not pregnant currently. Patient is not hospitalized and do not require supplemental oxygen. Patients received a deceased donor kidney transplant 3 years back, which make her a high risk patient. The patient's eGFR=61 ml/min. And the patient do not have any liver issues. And the patient is taking other medications that can be coordinated/given with Paxlovid. And the patient can also get/ has access to outpatient infusion or inpatient admission for remdesivir. & Paxlovid Dosing: Nirmatrelvir 300 mg 2x daily for 5 days and Ritonavir 100 mg 2x daily for 5 days & hard\\
    A patient, born on May 11, 1989, underwent a diagnostic test (NAAT) for Covid-19 and tested positive. The patient's weight stands at 40 kg and has a peculiar genetic blood disorder. A crucial indicator of kidney function, known as the glomerular filtration rate (GFR), is abnormally low, measuring at 32 ml/min, signifying impaired kidney function. And the patient's liver is functioning normally. Patient do not require increased oxygen. Presently, the patient refrains from the use of any additional medications, can get remdesivir. Given these intricate circumstances, determine the appropriate course of treatment for this patient. What treatment should the patient receive? & Paxlovid Dosing: Nirmatrelvir 150 mg 2x daily for 5 days and Ritonavir 100 mg 2x daily for 5 days & hard\\
    We are currently presented with a female patient who has recently undergone a Covid-19 antigen test, yielding a positive result. The patient does not need hospitalization. The patient is 18 years old and weighs 58 kg, not pregnant. Notably, the patient has received a liver transplant from a deceased donor. Now the patient's liver is very healthy. In terms of kidney function, the patient's glomerular filtration rate stands at 61 ml/min. Additionally, the patient is taking multiple medications, some of which may interact adversely with Paxlovid. However, the patient has physician's permission for remdesivir. & Remdesivir Dosing: 200 mg IV on day 1, 100 mg IV on days 2 \& 3 & hard\\
    We are currently presented with a female patient who has recently undergone a Covid-19 antigen test, yielding a positive result. The patient does not need increased oxygen. The patient is 18 years old and weighs 58 kg. Notably, the patient received a liver transplant from a deceased donor. In terms of kidney function, the patient's glomerular filtration rate stands at 21 ml/min. The patient does not have liver issues. Additionally, the patient is taking multiple medications, some of which may interact adversely with Paxlovid. However, the patient has been granted permission for the administration of remdesivir and have access to it. Given these intricate circumstances, What treatment should the patient receive? & Remdesivir Dosing: 200 mg IV on day 1, 100 mg IV on days 2 \& 3 & hard\\
    A patient has recently undergone NAAT test for Covid-19 and received a positive result. The patient is a female, not pregnant, and is only 12 years old, weighing 40 kg. The patient does not need increased oxygen. It is worth mentioning that the patient has a genetic blood disorder. Her glomerular filtration rate stands at 32 ml/min, indicating impaired kidney function. And the patient's liver is in a healthy state. Moreover, the patient is currently taking other medications that conflict with the usage of Paxlovid. Additionally, the patient does not have access for the administration of remdesivir. Given these complex circumstances, What treatment should the patient receive? & Outpatient treatment options not authorized or recommended. Place in monitoring and supportive care only & hard\\
    We are confronted with a patient who has yielded a positive result in a Covid-19 antigen test, signaling a true acute infection. But the patient does not need hospitalization. The patient is a female, not currently pregnant, and is in her 19th year, with a weight of 40 kg. Pertinently, the patient carries a hereditary blood disorder. The glomerular filtration rate is alarmingly low, measuring at 36 ml/min, highlighting impaired kidney function. Patient has always pay attention to her liver health, and her liver thanks her for that. Patient is taking drugs conflict with Paxlovid. Furthermore, the patient does not have access for the infusion center. & Molnupiravir dosing: 800 mg (four 200 mg capsules) orally twice daily for 5 days & hard\\
    We are faced with a patient who has obtained a true positive result with Covid-19 NAAT testing. The patient is not hospitalized. The patient is a female, currently not pregnant, and is 17 years old, with a weight of 40 kg. Notably, the patient is afflicted with a genetic blood disorder. A concerning aspect is the patient's glomerular filtration rate, which stands at 32 ml/min, indicating compromised kidney function. Patient have liver failures. Furthermore, the patient has not yet been granted authorization for the administration of remdesivir. & Outpatient treatment options not authorized or recommended. Place in monitoring and supportive care only & hard\\
    We are presented with a patient who has tested positive with Covid-19 NAAT. The patient does not require increased oxygen or hospitalization. The patient is a 19-year-old female weighing 42 kg and is not currently pregnant. It is noteworthy that the patient has undergone a surgical procedure involving the transplantation of healthy bone marrow stem cells to replace diseased or damaged bone marrow. The patient's glomerular filtration rate is measured at 29 ml/min, indicating compromised kidney function. Moreover, the patient does not have access to the use of remdesivir. & Molnupiravir dosing: 800 mg (four 200 mg capsules) orally twice daily for 5 days & hard\\
    We encountered a patient who has tested positive with Covid-19 NAAT. The patient is a 17-year-old female weighing 32 kg and is not currently pregnant. Patient does not need hospitalization. Importantly, the patient has been diagnosed with a debilitating lung disease characterized by airway inflammation and damage, leading to breathing difficulties. Additionally, the patient's glomerular filtration rate (GFR) is measured at 32 ml/min, indicating impaired kidney function. In the patient's liver examinations, there are some figures above or below the standard range. It is noteworthy that the patient has been granted permission for the administration of remdesivir and have access to it. & Remdesivir Dosing: 5 mg/kg IV on day 1 followed by 2.5 mg/ kg IV once daily from day 2 to day 3 & hard\\
    We are faced with a patient who has tested positive with Covid-19 NAAT. Patient does not require increased oxygen or hospitalization. The patient is an 11-year-old female weighing 32 kg and is not currently pregnant. Notably, the patient has been diagnosed with a challenging lung disease (bronchiectesis) characterized by airway inflammation and damage, resulting in breathing problems. Patient's liver is very healthy. Moreover, the patient has not yet been granted authorization for the use of remdesivir. & Outpatient treatment options not authorized or recommended. Place in monitoring and supportive care only & hard\\
    Patient has tested positive for covid-19 and requires supplemental oxygen. & Check CDC/IDSA/NIH Guidance & medium\\
    An 18-year-old patient tested positive for Covid-19, weighs 60 kg. He is asymptomatic. & Outpatient treatment options not authorized or recommended. Place in monitoring and supportive care only & medium\\
    An 18-year-old tested negative for Covid-19. He weights 60 kg. & Vaccination and booster is recommended & medium\\
    We have a patient who has a positive Covid-19 antigen test, female, 28 years old and weighs 65 kg. Patient does not need increased oxygen. Patient received a deceased donor kidney transplant 3 years back and has GFR=61 ml/min. She takes other medications that can be coordinated / given with Paxlovid. And the patient has been approved for remdesivir. & Paxlovid Dosing: Nirmatrelvir 300 mg 2x daily for 5 days and Ritonavir 100 mg 2x daily for 5 days & medium\\
    We have a 31-year-old thin patient testing positive for Covid-19 NAAT. She weighs 40 kg. She does not need increased oxygen. Patient has a history of genetic blood disorder. The patient has chronic kidney disease with eGFR=32 ml/min. She does not take other medications. And the patient has insurance's permission for remdesivir infusion so has access to it. & Paxlovid Dosing: Nirmatrelvir 150 mg 2x daily for 5 days and Ritonavir 100 mg 2x daily for 5 days & medium\\
    We have an 18-year-old female patient who has a positive antigen test, weighing 58 kg. Patient does not need hospitalization. Patient received a liver transplant from her brother 2 years back and it was very successful. The patient has a GFR=61 ml/min. And the patient takes other medications that conflicts with Paxlovid. She has her insurance's permission for remdesivir for outpatient infusion. & Remdesivir Dosing: 200 mg IV on day 1, 100 mg IV on days 2 \& 3 & medium\\
    We have a female patient who has a true positive result for Covid-19 antigen test, not pregnant, 17 years old and weighs 40 kg. She does not need hospitalization. Patient has a genetic blood disorder. Her GFR=29 ml/min. She took medications conflict with Paxlovid. And the patient has been granted permission for the infusion of remdesivir. & Remdesivir Dosing: 200 mg IV on day 1, 100 mg IV on days 2 \& 3 & medium\\
    We have a female patient testing positive for Covid-19 NAAT test. Patient does not need increased oxygen. She is 12 years old, not pregnant and weighs 41 kg. Patient has a genetic blood disorder and GFR=32 ml/min and does not have any liver problem. She takes medications that conflict with Paxlovid. She does not have permission for remdesivir. & Outpatient treatment options not authorized or recommended. Place in monitoring and supportive care only & medium\\
    We have a patient who has a positive Covid-19 antigen test, female, not pregnant, 18 years old and weighs 58 kg. Patient does not need hospitalization. Patient received a donated liver from her uncle and it was a successful transplant surgery. Her liver is functioning well. The patient has a GFR=61 ml/min. And the patient takes other medications that conflict and interact with Paxlovid. Patient does not have permission for remdesivir infusion. & Molnupiravir dosing: 800 mg (four 200 mg capsules) orally twice daily for 5 days & medium\\
    We have a patient who has a true result for Covid-19 NAAT, female, not pregnant, 17 years old and weighs 40 kg. She does not need hospitalization. Patient has a genetic blood disorder. The patient has a GFR=29 ml/min. And she does not yet have permission for remdesivir. & Outpatient treatment options not authorized or recommended. Place in monitoring and supportive care only & medium\\
    We have a patient with a positive Covid-19 NAAT, female, 19 years old and weighs 42 kg. Patient does not need hospitalization. Patient is not pregnant. Patient had a surgery involving replacing a person's diseased or damaged bone marrow with healthy bone marrow stem cells. The patient has a GFR=29 ml/min. And the patient does not yet have permission/authorization for remdesivir. What treatment should the patient receive? & Molnupiravir dosing: 800 mg (four 200 mg capsules) orally twice daily for 5 days & medium\\
    We have a 17-year-old female patient with a positive Covid-19 NAAT. Patient does not need hospitalization. She weighs 32 kg. Patient is not pregnant. Patients has a history of lung disease that makes it difficult to breathe due to inflammation and damage to the airways. The patient has a GFR=29 ml/min. And the patient received permission for remdesivir and have access to it. What treatment should the patient receive? & Remdesivir Dosing: 5 mg/kg IV on day 1 followed by 2.5 mg/ kg IV once daily from day 2 to day 3 & medium\\
    We have a patient with a positive NAAT Covid-19 test, female, 11 years old and weighs 32 kg. Patient does not need hospitalizations. Patient is not pregnant. Patient has history of chronic lung disease (bronchiectesis) that makes it difficult to breathe due to inflammation and damage to the airways. Patient's GFR is 4mL/min and does not take any drugs that cause conflict with Paxlovid. And the patient has not received authorization for remdesivir. & Outpatient treatment options not authorized or recommended. Place in monitoring and supportive care only & medium\\
    The patient has covid-19, oxygen saturation (Sp02) at room air is low and needs supplemental oxygen. & Check CDC/IDSA/NIH Guidance & easy\\
    A 34-year-old patient has tested positive for SARS-CoV-2 and weighs 62 kg. Patient does not need increased oxygen. He is asymptomatic and has no major clinical risk factors in his history. & Outpatient treatment options not authorized or recommended. Place in monitoring and supportive care only & easy\\
    A 16-year-old boy tested negative for Covid-19. His weight is 44 kg. & Vaccination and booster is recommended & easy\\
    A 28-year-old female has a past medical history of kidney transplant and takes immunosuppression drugs. She weights 65 kg and has tested positive for Covid-19. She does not need hospitalization. And she does not have any chronic kidney disease with a GFR of 94 mL/min. Her immunosuppressive medications do not interact with paxlovid and she can hold few other home medicines while taking paxlovid. She can also take remdesivir at nearest infusion center. & Paxlovid Dosing: Nirmatrelvir 300 mg 2x daily for 5 days and Ritonavir 100 mg 2x daily for 5 days & easy\\
    A 31-year-old female, weighing 40 kg, tested positive for covid-19. The patient does not need hospitalization. She has a history of genetic blood disorder and cardiovascular disease. She has a GFR of 32 mL/min and does not have any hepatic impairment. But does not take any medications. She has access to outpatient infusion or inpatient admission for remdesivir. & Paxlovid Dosing: Nirmatrelvir 150 mg 2x daily for 5 days and Ritonavir 100 mg 2x daily for 5 days & easy\\
    We have a patient who is covid-19 positive, female, 18 years old and weighs 58 kg. Patient is taking immunosuppressive drugs. Patient is not hospitalized. The patient has a GFR=61 ml/min. Patient does not have hepatic impairment. The patient is taking other medications that cannot be held, dose adjusted or substituted while Paxlovid is given. The patient has access to outpatient infusion or inpatient admission for administration of remdesivir. & Remdesivir Dosing: 200 mg IV on day 1, 100 mg IV on days 2 \& 3 & easy\\
    We have an 18-year-old non pregnant female patient who has a positive covid-19 test and weighs 58 kg. Patient is not hospitalized. The patient is a bone marrow transplant recipient at the age of 8 years. The patient has a GFR=31 ml/min. But the patient has hepatic impairment. She takes other medications that cannot be dose-adjusted with Paxlovid. Patient can get remdesivir. & Remdesivir Dosing: 200 mg IV on day 1, 100 mg IV on days 2 \& 3 & easy\\
    We have a 12-year-old non pregnant female weighing 40 kg testing positive for Covid-19. The patient does not need increased oxygen. She has chronic kidney disease and is at high risk for disease progression due to Covid-19. Patient has a genetic blood disorder and GFR=32 ml/min. Patient does not have any liver issues. She takes medications that conflict with Paxlovid. And the patient does not have access to remdesivir. & Outpatient treatment options not authorized or recommended. Place in monitoring and supportive care only & easy\\
    We have a patient who has a positive result for covid-19, female, not pregnant, 18 years old and weighs 40 kg. Patient does not require increased oxygen. The patient is at high risk for Covid-19 disease progression. She has GFR of 36 mL/min, does not have hepatic impairment. The patient is taking other drug conflict with Paxlovid. And the patient does not has authorization for outpatient infusion for remdesivir. & Molnupiravir dosing: 800 mg (four 200 mg capsules) orally twice daily for 5 days & easy\\
    A 17-year-old non pregnant female weighs 40 kg and tested positive for Covid-19. Patient does not need hospitalization. The patient is at high risk for clinical deterioration. The patient has severe renal impairment. And she does not yet have authorization for administration of remdesivir. & Outpatient treatment options not authorized or recommended. Place in monitoring and supportive care only & easy\\
    We have a patient with a positive covid-19 test, female, 19 years old and weighs 42 kg. Patient is not hospitalized. Patient is not pregnant. Patients had a recent surgery which puts her at high risk for clinical decompensation from Covid-19. The patient has a GFR=29 ml/min. And she does not yet have access for administration for remdesivir. & Molnupiravir dosing: 800 mg (four 200 mg capsules) orally twice daily for 5 days & easy\\
    We have a patient with a positive covid-19 test, female, 17 years old and weighs 32 kg. Patient does not need increased oxygen. Patient is not pregnant. Patient has a history of chronic lung disease which puts him at high risk. The patient also has renal impairment. And the patient has access for administration for remdesivir. & Remdesivir Dosing: 5 mg/kg IV on day 1 followed by 2.5 mg/ kg IV once daily from day 2 to day 3 & easy\\
    We have a patient with a positive covid-19 test, female, 11 years old and weighs 32 kg. Patient is not hospitalized. Patient is not pregnant. Patient is at high risk for decompensation from Covid-19. And the patient has not received authorization for remdesivir. & Outpatient treatment options not authorized or recommended. Place in monitoring and supportive care only & easy\\
  \hline[1pt,black]
\end{longtblr}


\end{document}